\documentclass[lettersize,journal]{IEEEtran}
\usepackage[utf8]{inputenc}
\usepackage{amsmath,amsfonts}
\usepackage{algorithmic}
\usepackage{algorithm}
\usepackage{array}
\usepackage[caption=false,font=normalsize,labelfont=sf,textfont=sf]{subfig}
\usepackage{textcomp}
\usepackage{stfloats}
\usepackage{url}
\usepackage{verbatim}
\usepackage{graphicx}
\usepackage{cite}
\usepackage{multirow}
\begin{document}

\title{Transferable Spatial Temporal
Coherence Adversarial Attack on
Black-Box Vision Language Models for
Autonomous Driving}

\author{Heyam M. Bin Jahlan, Areej M. Alhothali, Abeer Alhothali}

\markboth{}%
{Shell \MakeLowercase{\textit{et al.}}: A Sample Article Using IEEEtran.cls for IEEE Journals}

\IEEEpubid{}

\maketitle

\begin{abstract}
The rapid integration of Vision Language Models (VLMs) into sensitive systems such as autonomous driving systems introduces critical safety vulnerabilities that remain largely unexplored in exist studies. While adversarial attack robustness has been extensively studied for image-based models, the susceptibility of VLMs to temporally-aware adversarial attacks against video in driving context poses a distinct and under examined threat. In this paper, we introduce novel adversarial attack framework against video targeting VLM models used for autonomous driving scenes named \textbf{Spatial Temporal Coherence Adversarial Attack (STCA)}. Our attack comprise from three primary stages: modalities expansion, Spatial attack, and STCA attack. In modalities expansion, we extract multi-caption of each video and propose caption-guided frame selection method in order to ensure that adversarial perturbation target the most semantically significant frames. Secondly.In spatial attack, we apply YOLO-guided object mask in vital frames to craft effective perturbation and preserve high similarity. Then the perturbed video generated fed into STCA stage that exploits LanguageBind model  as surrogate model to disrupt cross-frame temporal coherence using motion guided mask.Our method operate under black box threat model against victim target VLMs, relying solely on transferability from white-box surrogate model.We conduct our experiments on the BDD100K and nuScenes autonomous driving datasets across three VLM models: Video LLaVA-7B, Qwen2.5-VL-7B, and Dolphin. Experimental results demonstrate spatial attack achieves an attack success rate (ASR) 32.6 \%  with high structural similarity index (SSIM=0.933) while in our proposed STCA attack, significantly enhances ASR to 71\% against Video LLaVA -7B model. We gained strong transferability to Qwen2.5-VL-7B ASR= 45\% in spatial attack and 84.2\% in our STCA attack. Dolphin dataset deceived by 46.2\% in BDD100K dataset. While in nuSences dataset, we achieve in spatial attack only 57.6\% , 64.7\%, and and 37.6\% , after SCTA stage, we gain 83\%, 96.5\%, and 47.1\% in video-LLaVa, Qwen2.5, and Dolphins models respectively.  Semantic divergence is calculated using word overlap similarity. Our finding reveal that existing video language model, remain highly susceptible to adversarial attack in autonomous driving scenarios, underscoring the urgent need for robust defense for VLM models.  
\end{abstract}


\section{Introduction}
\begin{figure}
    \centering
    \includegraphics[width=1\linewidth]{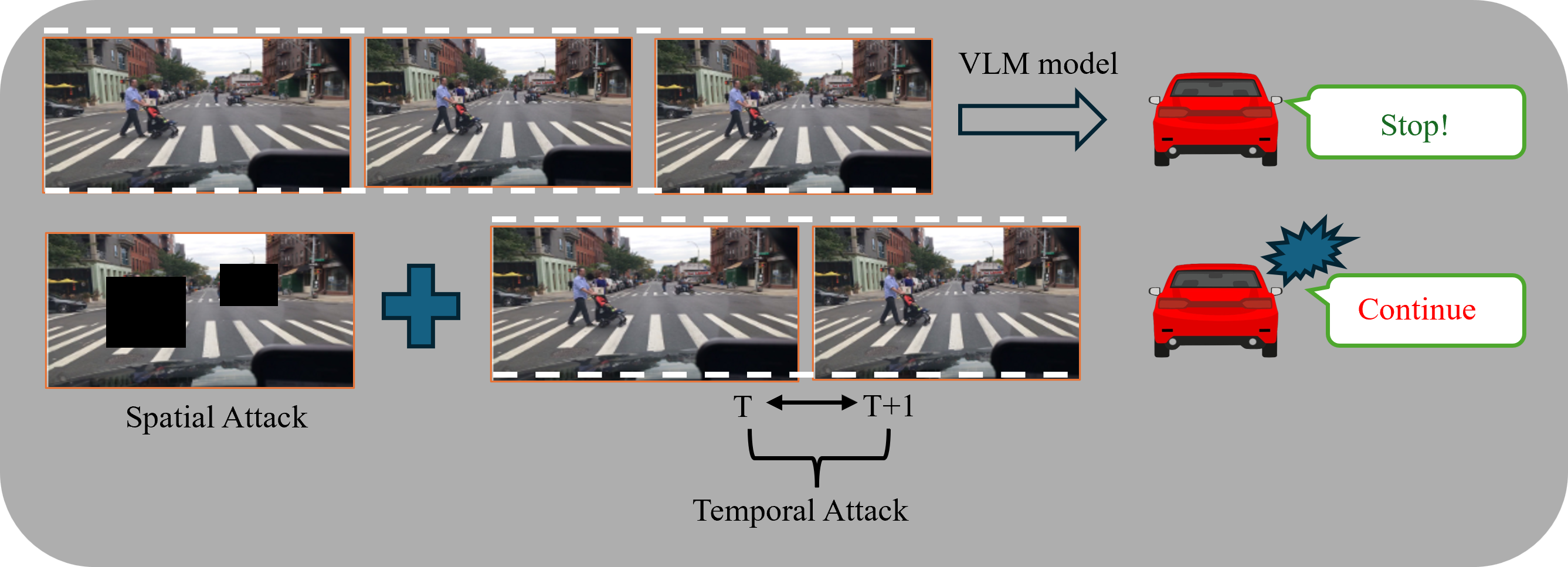}
    \caption{Enter Caption}
    \label{fig:intro}
\end{figure}

\IEEEPARstart{T}{he} rapid progress of Vision Language Pre-training models (VLP) has transformed the field of multi-modal artificial intelligence, enabling systems to concurrently understand and analysis textual and visual information. Current models such as ALBEF ~\cite{li2021albeef}, CLIP ~ \cite{radford2021clip}, and TCL ~\cite{yang2022tcl} have demonstrated exceptional performance across a wide range of downstream tasks including image captioning, visual question answering, and image-text retrieval. These models leverage extensive contrastive learning objectives in order to align textual and visual representation in a shared embedding space achieving unprecedented level of cross-modal understanding. However, despite their remarkable capabilities, VLM model exhibit inherent susceptibility to adversarial attacks, where the carefully designed imperceptible perturbations to the input can significantly degrade model performance, raising substantial concerns regarding their reliability and security in practical applications. \\
Adversarial Attacks on deep learning models have been extensively examined in the uni-model context, specially for image classification FGSM ~\cite{goodfellow2015fgsm} and PGD~\cite{madry2018pgd}. The extension of adversarial attacks to the multi-modal domain, however, current unique challenges that are not fully addressed by existing uni-modal techniques. In the case of VLM model, adversarial perturbations deceive both the cross-modal alignment mechanism and the visual encoder, making the attack issue significantly more complex. Previous methods in adversarial attack for VLM models such as AdvCLIP ~\cite{zhang2022advclip} have shown the feasibility of generating adversarial examples that can deceive these models in \textbf{white-box} scenario, where the attacker has full knowledge about the model parameters. Nevertheless, the transferability of such attacks across different VLM architectures remain largely underexplored that create substantial performance gap between black-box and white-box scenarios. \\
While image-based adversarial attack has been extensively examined, the primary limitation of current adversarial attacks studies for VLM models is the failure to evaluate the temporal dynamics inherent in video data. The issue of generating adversarial perturbation against temporal consistent of video sequence remain an open research challenge. This additional dimension of complexity guides us to disrupt the temporal coherence between each two consecutive frames to capture the movements regions instead of only fool the model on individual frame. The temporal aspect is critical in autonomous driving where the vehicle must be aware about the relationship between object during video. Temporal attack is not addressed before in existing studies that only focus in image spatial attack for individual frame and neglected temporal structure resulting in a non-comprehensive evaluate for attack effectiveness.\\
Additionally, existing adversarial attack methods in VLM models that targeting the consistency between visual and textual modalities typically apply perturbation uniformly across the entire image. on the contrary, our proposed method aimed to consider the semantic relevance regions in different frame and neglected regions that contribute minimally to deceiving the cross-modal alignment mechanism such as background and sky. Advances techniques of object detection and attention-based in video that identify semantically meaningful regions, remain largely unexplored in adversarial attack pipelines for VLM models.\\
To address these limitations, we propose \textbf{Spatio-Temporal Coherence Adversarial Attack STCA} framework for VLM model in video data. Our proposed framework innovate three key stages that collectively enhances the effectiveness of adversarial attack on video based VLM models. \textbf{First}, we extract multi-captions for each video using two different VLM model in order to capture the diversity of textual prompt, after that, we propose \textbf{Caption Guided Frame Selection} strategy that utilize CLIP model to select the most semantically relevant frames from the sequence of video with respect to associated description. To ensure extracting most informative frame, we compute the cosine similarity between the visual embeddings of each frame and the textual embeddings of the associated captions, the top $K$ frames that maximize the semantic alignment have been selected. This strategy ensure that, the adversarial perturbation is concentrate to only the informative frame while maximizing the similarity between the original and the adversarial videos. \\
Rather than applying perturbation uniformly across entire frame, we utilize masking strategy for each critical frame based on YOLOv8 object detector ~\cite{yolov8} that constrains adversarial perturbation to semantically meaningful regions and restrict the adversarial perturbation to these regions. This mask-guided perturbation strategy ensure that we disrupt the cross-modal alignment and enhancing both the effectiveness of the attack and the imperceptibility of the perturbation to human observe.\\
We also extend the adversarial attack framework to disrupt the coherence of temporal domain that explicitly perturbs the movement regions in the video by taking the difference between each two adjacent frames. Our STCA attack leverages LanguageBind video encoder ~\cite{zhu2024languagebind} to generate adversarial video by calculate frame-level visual feature and optimize adversarial perturbations that maximize the deceptive of temporal relationship. The loss function of STCA attack is aimed to minimize the cosine similarity between features of consecutive frames weighted by motion mask ensuring that, the adversarial perturbations target regions of high temporal dynamics as illustrated in figure \ref{fig:intro} .\\

We conduct our comprehensive experiments on the BDD100K~\cite{yu2020bdd100k} and nuScenes ~\cite{caesar2020nuscenes} datasets across three divers VLMs: Video LLaVA-7B ~\cite{lin2024videollava}, Qwen2.5-VL-7B ~\cite{qwen25vl2025}, and Dolphin ~\cite{ma2024dolphins}. We adopt Jaccard word-overlap similarity to ensure model rigorous. Our experimental results demonstrate the effectiveness of our attack against these three models on both datasets.  \\
The principal contributions of our framework summarized as follows:
\begin{itemize}
    \item We propose three-stage adversarial attack framework that combines modalities expansion, spatial, and temporally-aware cross-modal perturbation designed to target VLM models in autonomous driving systems. 
    \item We conduct our robustness evaluation of driving-specefic VLMs on the BDD100K dataset contain 1000 driving video and nuScenes datasets with 85 scenes against three divers target model. 
    \item We demonstrate that adversarial perturbation transfer effectively across divers VLMs models under the black-box threat that poses serious security implications. 
    \item We observe that, driving-specific VLMs such as Dolphin exhibit comparatively more robustness than general-purpose model.
    \item Our finding highlights the urgent need adversarial defense mechanism in autonomous driving pipelines. 
\end{itemize}

The rest of our paper introduces related work in third section, problem and motivation in the fourth section. In the fifth section, we introduce our proposed framework in detail. Finally, our finding results and discussion  is presented in section 6.

\section{Related Work}
\subsection{Adversarial Attack against VLM}
Recently, the rapid proliferation of VLM in real-world application lead up to the extension of adversarial attack from uni-model systems to multi-model. Zhang et al. ~\cite{zhang2022advclip} was the first paper observe that, by perturbing each modality separately, the attack will be degrade due to the contradiction between different modalities. They propose novel multi-modal attack called collaborative multimodal attack (co-attack) which collectively attack both image and text simultaneously on two VLM architectural types Fused VLM and aligned VLM and three V+L tasks, visual entailment, visual grounding, and image-text retrieval. They approve the VLM is susceptible to adversarial attack, especially that crafted to perturb image and text representation in joint embedding space. Zhou el al. ~\cite{zhou2023advclip} extend this work by construct universal adversarial patch targeting all downstream tasks of victim cross-modal pretrained encoder such as CLIP model. Their AdvCLIP approach that downstream-agnostic adversarial example, could be built without target specific task a model would be deployed on. However, their empirical evaluation to classification and retrieval tasks, leaving other tasks unexplored. \\

Yin et al ~\cite{yin2023vlattack} have been explored the adversarial attack in black-box setting by propose Vision Language Attack Strategy (VLattack) made up of two primary levels: single-modal level, and multi-modal level. They observed that models fine-tuned in specific task inherit the adversarial vulnerabilities of their pre-trained visual encoder. Zhao et al. ~\cite{zhao2023robustness} carried out a large scale evaluation of adversarial across several large VLM, they discovered all powerful tested VLM models remain vulnerable to adversarial attack. Recently, Zhang el al. ~\cite{zhang2025anyattack} proposed AnyAttack that leverage the clean image to be transferred to attack other different VLMs and generating any desired output by training their generator on LAION 400M dataset. While Lu et al.~\cite{lu2023set} developed set-level Guidance attack (SGA) that enhance transferability by generating adversarial perturbation across divers cross-modal interaction of multiple text-image pairs. They showed the stronger transferability when attacked all the modalities simultaneously. in case of diffusion model leveraged by Guo et al.~\cite{guo2024efficient} in order to generate transferable adversarial example against VLM. Their attack achieve robust performance and strong quality. Unlike previous existing methods that suffer from high computational cost due to complex structure and number of iterations, their AdvDiffVlm utilize Adaptive Ensample Gradient Estimation (AEGE) to modify the score function thereby embedding adversarial semantic that enhance transferability. Recent work by Liu et al. ~\cite{liu2025evaluating} introduce maximizing Information Entropy (MIE) to enhance attack efficacy by directly inducing layer by layer perturbation in feature representation. They observed internal feature perturbation is more effective and comprehensive than approaches that only rely on final output layer. 
\subsection{Adversarial Attack against Video model}
The extension of static image to video model is arising from the temporal dimension, As each frame in video can be treated as a naive application of image level attack. Wei et al. ~\cite{wei2022boosting} targeted set of temporal translated video to apply adversarial perturbation on them. To achieve better transferability than image attack method, they ovoid overfitting to the white box model. Due to the high computational cost, subsequently, they aimed in ~\cite{wei2023adaptive} to attack video CNN and ViT models by generate adversarial video from white-box image model named Image To Video Attack (I2V). Their approach here easy to perform and achieves better performance without training video model. In their further work~\cite{wei2023efficient}, They assess the robustness of video recognition model spatial and temporal focus attack which perturb only temporally salient regions instead of disturbing all frames uniformly that reduce attack query numbers. Kim et al. ~\cite{kim2023breaking} exploit image models and image data in their Universal Adversarial Perturbation (UAP) framework to attack video models using image-level surrogate model. Additionally, they proposed Breaking Temporal Consistency (BTC) to attack temporal aspect of video by reducing the features similarity between consecutive frames and by reducing the similarity between UAPs (temporal similarity loss).\\
whereas the current studies of video adversarial attack target action recognition model, we target video language model performing scene description of autonomous driving scenario that demand to measure success by semantic divergence in natural language output not by classification accuracy on fixed label set.   

\subsection{Adversarial Attack against Autonomous Driving }
The intersection of autonomous driving and adversarial robustness has garnered growing attention due to the high-stakes implications of perception error in this field. Early research in this area focused on attacking object detector. Several studies showed that adversarially designed patches applied to road surface or vehicle could cause detection system to misclassify object in their range of view. Chung et al.~\cite{chung2024typography} in their Typography-based attack are among the first emphasize practical transferability in traffic scenes, relying on decision-making capabilities of VLM. They propose dataset-agnostic framework for automatically generating false answer and linguistic augmentation to misdirect image-level and region-level reasoning in realistic road context. \\ 
PG-Attack, proposed by Fu et al.~\cite{fu2024pgattack} similarly targets deployed vision foundation models with a black-box that combines precision mask perturbation PMP with deceptive text patches (DTP) techniques, and their results suggest strong empirical defense across many VLM models. Zhang et al. ~\cite{zhang2026advlm} explicitly framed as the first attack tailored to VLM in autonomous driving, thye proposed ADvLM which marks as a sharper methodological step identified AD-specific difficulties: various prompt paraphrased but semantically equivalent and time-series scene with multiple viewpoint. Their scenario associated frame selection and semantic invariant prompt set it apart from typography and PG-Attack which are more transfer- and patch-oriented and less clearly constructed around temporal driving semantics. \\
Recently, studies broaden both the attack surface and autonomy setting, CAD ~\cite{wang2025black} is positioned directly against ADvLM, they contends that, the field has only a white-box AD-specific attack and present the first black-box attack tailored for AD VLM using decision-chain disruption and risky scene induction. There is an clear chronological relationship, because CAD extends the similar safety-critical setting while easing attacker access and displaying actual route-completion degradation. The goal of trustworthy-AV is more than attack proposal ~\cite{fime2025trustworthy}; it is a robustness benchmark among vision encoders under targeted and untargeted adversarial conditions with sim-CLIP exhibiting the best model-dependent resilience. Its limitation is that it evaluates encoder robustness instead of the entire decision pipeline studied by CAD or ADvLM. \\
Natural reflection~\cite{liu2025natural} opens different line by switching from test time perturbation to training time poisoning utilizing faint reflection triggers to cause long replies and detrimental inference delay while maintaining clean-input performance. Its specialization is main limitation; it enhances rather than replace perturbation-based attack studies since it focuses on delayed response behavior rather than general semantic misdriving. Robotic VLA attack expand the realm beyond driving by demonstrating that jailbreaking-style textual attack can gain persistent control authority over vision language action  ~\cite{jones2025vla}. Then ADVEDM ~\cite{wang2025advedm} pushes embodied attack toward fine-grained semantic manipulation claiming that while selective object addition or removal preserve context and induce valid but incorrect actions in both driving and manipulation tasks, many previous attacks break too much scene semantic and thus produce invalid outputs. Two keys issues are combined in most recent studies. First, robustness varies significantly by architecture, LLaVA is much easier to attack under BIM, PGD, and spectral attack than Qwen2.5-VL in nondriving agent scenario ~\cite{paredes2026modern}. Second, architecture diversity alone is insufficient to address this issue in autonomous driving, because physical patches transfer strongly between Dolphins, OmniDrive, and LeapVAD especially when model share a CLIP-based encoder ~\cite{fernandez2026transferability}. \\
The clearest bridge in these concern is is UCA ~\cite{kong2025uca}, It is specifically criticizes previous AD attacks as being primarily logit-level, and then suggests a physically achievable feature-space camouflag attack that is independent of user text and robust to viewpoint and scale changes.\\
The main unresolved limitation across the literature, there is no study address all of these in addition to temporal aspect at once, so current evidence is strong on isolated threat models but still fragmented across temporal dynamics, attacker access, and end-to-end control sequence. As the best of our knowledge, this is the first work to address the temporal dimension of adversarial attack in video language models in autonomous driving context.

\section{Problem and Motivation}

Most existing adversarial attacks against VLM in autonomous driving focus primarily on static image perturbation. Our method aimed to deceive the model to produce incorrect decisions by carefully designing unnoticeable perturbations into the input data. Specifically, an attacker applies adversarial perturbations on a benign video rather than a static image. Formally, let a driving video sequence be represented as: $V={f_1,f_2,\dots ,f_T}$
where $f_T$ donate the $T_{th}$ frame. A victim VLM model process the video and generate a textual response: $y= \mathcal{F}(V,Q)$. where $(Q)$ represents a textual query or prompt, and $(\mathcal{F})$ donate the multimodal inference function. VLM victim model donated as $F_0$ that induced to output undesirable response $y^*$ under adversarial perturbed input, $y^*=\mathcal{P}(V,Q)$ where $\mathcal{P}(.)$ denotes the adversarial perturbation function. Formally, this manipulation is defined by maximizing the likelihood of $y^*$ response under adversarial input.

\begin{equation}
\max_{P} \log P\left(y^* \mid \mathcal{P}(V, Q)\right)
\end{equation}

where ${P}$ is a probability function to obtain undesirable response $y^*$ when given a perturbed input.  $ F_0: Q \xrightarrow{} R$, where $R$ represents the response domain. Additionally, Existing studies have overlook temporal interaction between consecutive frames. In autonomous driving, semantic understanding is not determined solely by isolated spatial content, but rather by temporal continuity, object motion, causal ordering, and scene evolution over time. Consequently, subtle perturbation will induce temporal reasoning failure. To characterize this issue, we define temporal semantic consistency as the difference between consecutive masked frames across time.

\begin{equation}
\mathcal{C}(V)=\frac{1}{T-1}\sum_{t=1}^{T-1}\cos\big(\Phi(f_t) - \Phi(f_{t+1}))
\end{equation} 

where $\Phi(\cdot)$ donate the joint visual-semantic embedding function and the difference between $\cos(\cdot)$ detect similarity of motion between each two consecutive frames. To best temporal understanding, model should maintains high consisitency under nature temporal transitions. For example the model may incorrectly infer that a pedestrian crossed after a vehivle stopped rather than before it stopped resulting in dangerous decision inconsistencies.    
The key challenge lies in refinement perturbations that can subtly alter the model’s perception, while maintaining high transferability and frames similarity under diverse conditions. Specifically, the adversarial examples must induce models' decisions without compromising spatial and temporal consistency or raising human suspicion. \\

\begin{figure}
    \centering
    \includegraphics[width=0.5\linewidth]{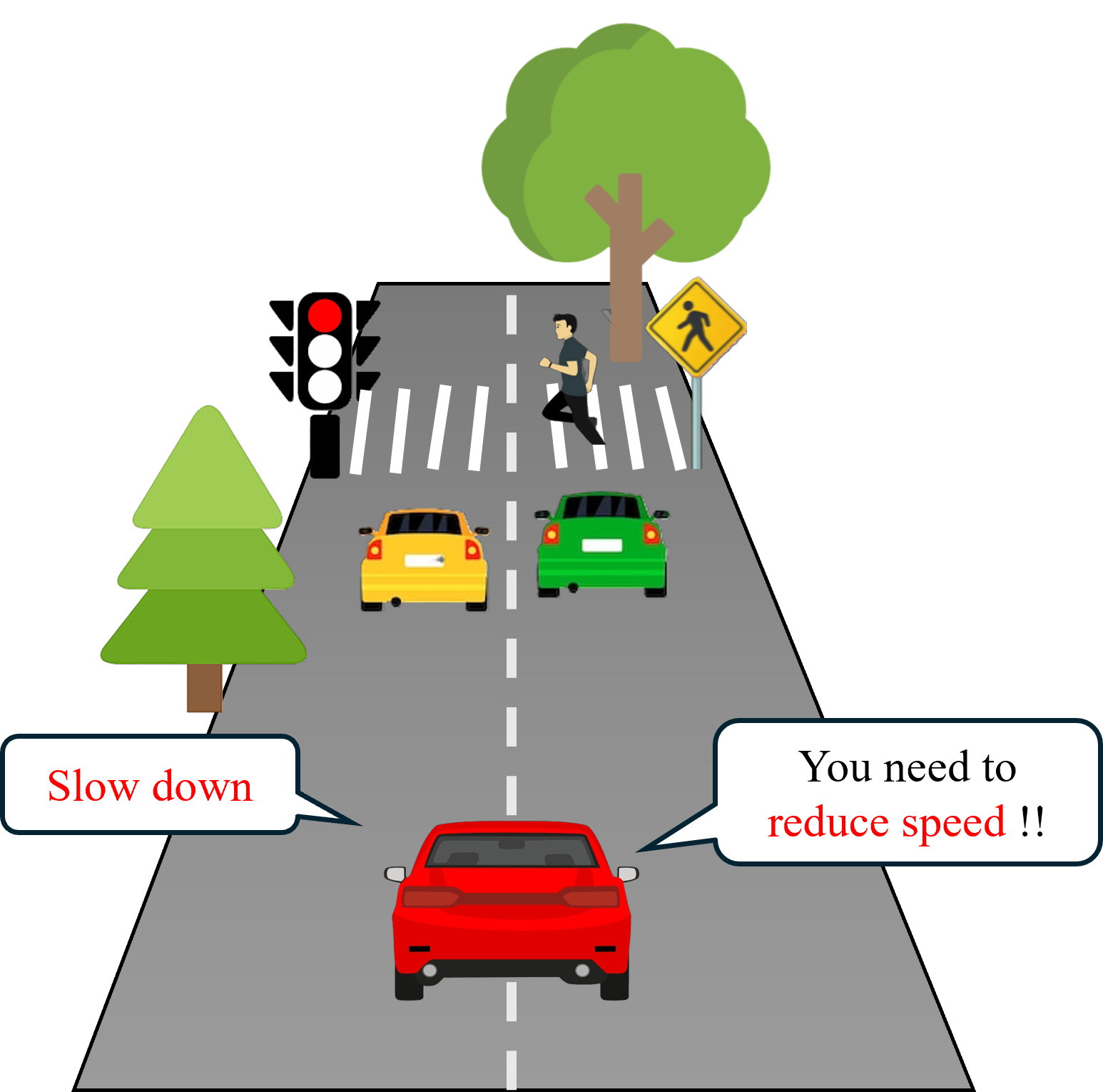}
    \caption{Diversity of textual prompts}
    \label{fig:motivation1}
\end{figure}

\begin{figure}
    \centering
    \includegraphics[width=0.5\linewidth]{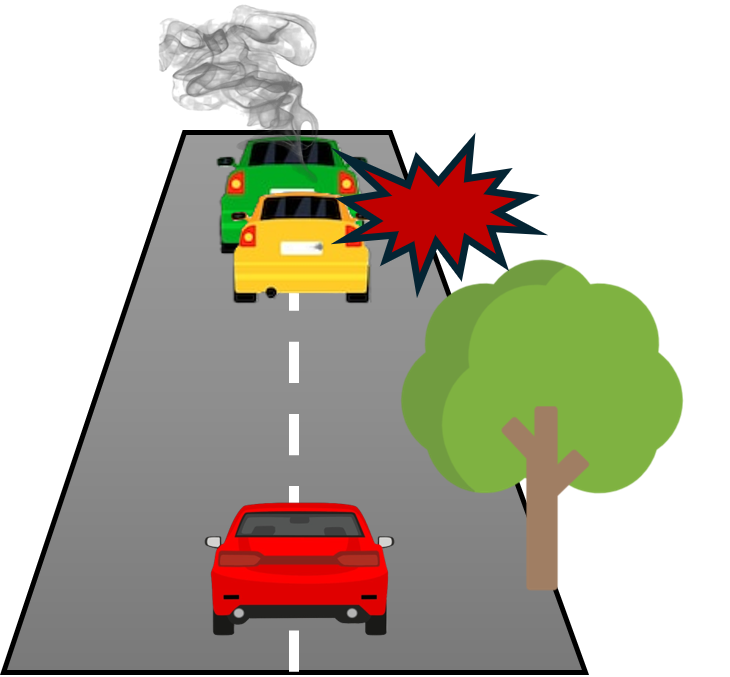}
    \caption{Diversity of visual scenario}
    \label{fig:motivation2}
\end{figure}

Furthermore, autonomous driving systems induce unique challenges that differ this attack from other general VLM attacks ~\cite{zhang2026advlm} can summrize these challenges as shown in figure \ref{fig:motivation1} and \ref{fig:motivation2}. Firstly, \textbf{the diversity of textual instructions}, one driving instruction can be expressed with different phrases, such as "slow down" and "reduce speed". This mean that, these instructions are written in different expressions but convey the same intent and semantics. second challenge represented in figure \ref{fig:motivation2} \textbf{the nature of the visual scenario} due to the perspective will be changed through driving.\\
In our method, we will design universal perturbation $\delta$ that ensures a stable, efficient attack across different instructions that equivalent semantics using various VLM models to candidate the best instructions that match with frames of video. We need to design more robust perturbation that adopt visual changes and temporal dependencies. In our method, the most vital frames that represent the change of the time will be selected based in the below equations.  
\begin{equation}
\mathcal{S} = \{f_t \mid \mathrm{Sim}(f_t,C) > \tau\} \
\end{equation}  

\begin{equation}
\Tilde{F}_t= \begin{cases} f_t+ \delta_t, & f_t \in \mathcal{S} \\ f_t , &\text {otherwise} \end{cases}
\end{equation} 

\begin{equation}
\tilde{V} = \mathcal{P}(\tilde{F})
\end{equation} 

Let $F_t$ denote the sequence of frames and $C$ represent a set of candidate captions generated for the entire video that identify semantically important frames within the video. the selected frame subset is defined in equation (4) where $\mathrm{Sim}(f_t,C)$ measure the semantic relevance between frame and caption set, while $\tau$ denotes a selection threshold. Frames that exhibit strong semantic alignment with the captions are considered temporally important and are therefore selected for adversarial manipulation. After frame selection, the perturbations are applied only to the selected frame as shown in equation (5), where $\delta_t$ represent adversarial perturbation added to the selected frame. Finally, the perturbed frame sequence is utilized to apply a temporal attack. \\

To address these gaps, our motivation stems from three key observations:
\begin{enumerate}
    \item Existing VLMs are predominantly trained on spatial-text alignment objectives rather than temporal causal reasoning.
    \item To design robust perturbations that induce spatial, temporal, and textual relations in the video.
    \item To ensure that our STCA attack exhibit stronger transferability across various VLM architectures
\end{enumerate}
Therefore, the motivation of our work, to exploit and reveal the temporal vulnerability of VLMs in autonomous driving systems. By targeting temporal semantic reasoning instead of relying solely on spatial and textual perturbations.

\section{Methodology}
\begin{figure*}[!t]
    \centering
    \includegraphics[width=0.8\textwidth]{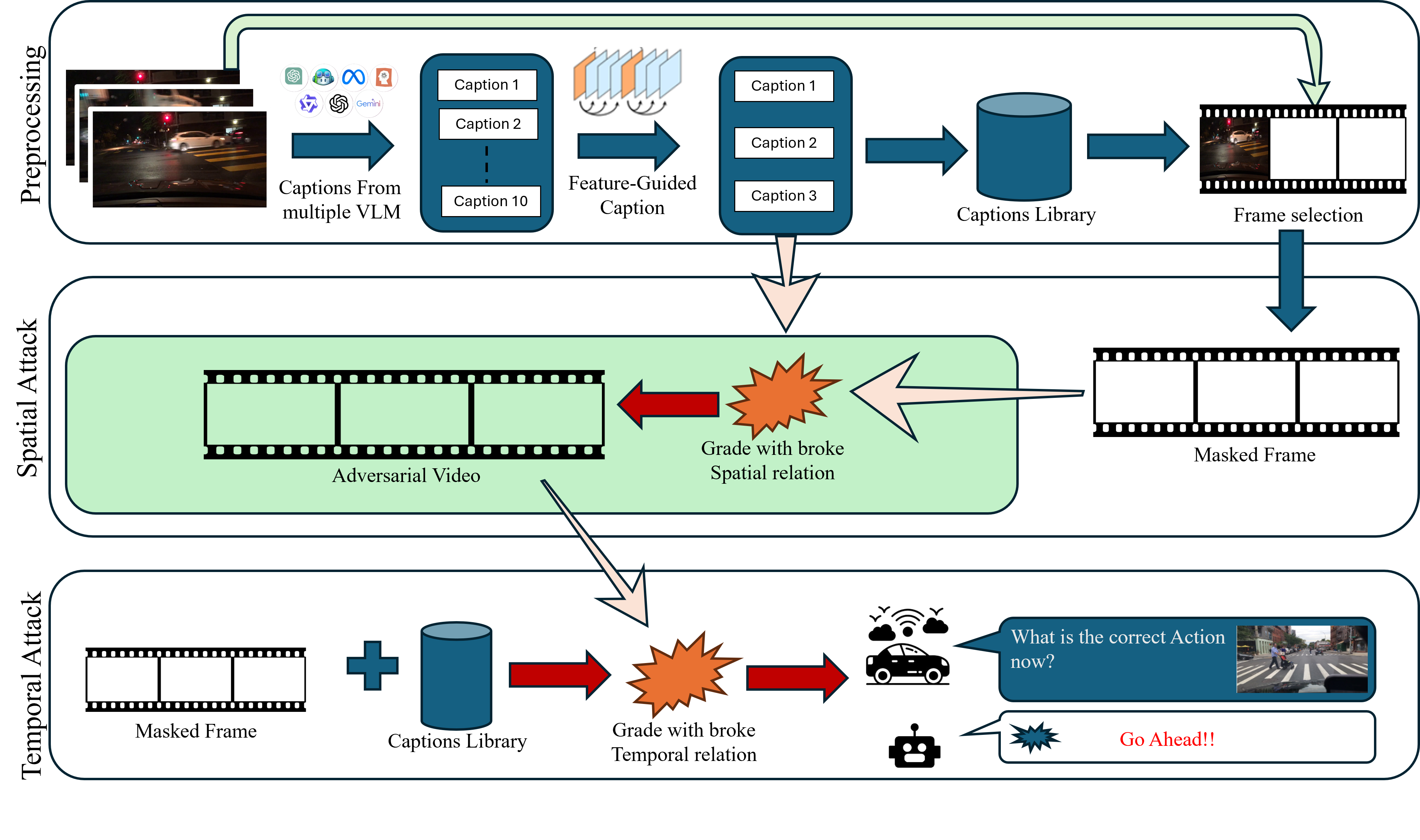}
    \caption{Our proposed framework consisting from three main stages: Modalities expansion, Spatial attack and temporal attack.}
    \label{framework}
\end{figure*}

To address the above challenges, we propose STCA Temporal Coherence  Adversarial attack which exploits spatial and temporal modalities using three phases as illustrated in figure \ref{framework}: modalities expansion, spatial attack, and temporal attack. In phase 1, the initial dataset prepared and fed it into Gemini and Video-LLAVA to extract multi-captions for each video. After that, the best related captions will be candidate using CLIP model. For spatial phase, we employ CLIP model to extract critical frames and perspectives to apply perturbation for them. Finally, in a temporal attack, we trained LanguageBlind model to disrupt temporal reasoning consistency and transferred it to other models. The whole description of STCA adversarial attack is illustrated in Algorithm \ref{alg:method}.

\begin{algorithm}[H]
\caption{STCA Adversarial Attack}\label{alg:method}
\begin{algorithmic}
\STATE 
\REQUIRE Dataset $\mathcal{D}$, video frames $v=\{f_1,f_2\dots f_t\}$, CLIP Vision Encoder $E_v$ , CLIP Textual Encoder $E_t$
\ENSURE Adversarial Video\\
\STATE \textbf{//Selected captions for each video}
\FOR{each video $v_i \in \mathcal{D}$}

 \STATE $C^{G} \xleftarrow{} \{c^{G}_1 , c^{G}_2 \dots c^{G}_k$
 \STATE $C^{L} \xleftarrow{} \{c^{L}_1 , c^{L}_2 \dots c^{L}_k$
 \STATE $C \xleftarrow{} C^{G} \cup C^{L}$
 \STATE \FOR{each caption $c_k \in {C}$}
 \STATE \FOR{each frame $f_t \in {v_i}$}
 \STATE $v_t \xleftarrow{} E_v (f_t)$
 \STATE $u_k \xleftarrow{} E_t (c_k)$
 \STATE \textbf{\emph{//Compute cosine similarity}}
 \STATE $s_{t,k} = \cos(v_t,u_k)$
\ENDFOR
\STATE \textbf{\emph{//Compute caption score}}
\STATE $S(c_k, v_i) = \frac{1}{T} \sum_{t=1}^T s_{t,k}$
\ENDFOR
\STATE \textbf{\emph{//Select top-ranked caption}}
\STATE $C^{*} =\operatorname{TopM}\big(c_k,v_i\big)$
\ENDFOR
\STATE \textbf{\emph{// Caption-guided frame selection}}
\FOR{each video $v_i \in \mathcal{D}$}
\STATE \FOR{each caption $t_j \in {C^*}$}
\STATE $\mathbf{u}_j \leftarrow \dfrac{E_t(t_j)}{\|E_t(t_j)\|_2}$

\ENDFOR
\STATE $\bar{\mathbf{u}} \leftarrow \dfrac{\sum_{j=1}^3 \mathbf{u}_j}{\left \|\sum_{j=1^3} \mathbf{u}_j \right \|_2}$
\STATE \FOR{each frame $f_i \in {v_i}$}
\STATE $\mathbf{v}_i \leftarrow \dfrac{E_v(f_i)}{\|E_v(f_i)\|_2} $
\STATE $s_i \leftarrow \mathbf{v}_i^\top \bar{\mathbf{u}} $
\ENDFOR
\STATE $\mathcal{F}^* \leftarrow \underset {\mathcal{S} \subseteq \{1,\dots ,N\},| \mathcal{S}|=k}{\arg \max} \sum_{i \in \mathcal{S}}s_i $
\ENDFOR
\FOR{each video $v_i \in \mathcal{D}$}
\STATE \FOR{each frame $F_j \in {\mathcal{F}^*}$}
\STATE $\mathcal{M} \leftarrow YOLOv8 \text{  } detector(F_j)$
\STATE $i^{adv} \leftarrow \text{Clip}_\epsilon \! \left (f_j\cdot \mathcal{M}_i (i^{adv}+\alpha \cdot \text{sign}(\mathbf{g}_k)) \cdot (1-\mathcal{M}_i\right) $
\STATE\textbf{\emph{// Temporal Coherence Attack }}
\STATE $\mathcal{M}_{motion} = \mathcal{M}_{j} - \mathcal{M}_{j+1}$
\STATE $\delta_t \leftarrow \alpha \cdot \text{sign}(\mathbf{g}_k)\cdot \mathcal{M}_{motion}$
\STATE $i_t^{adv} \leftarrow \text{Clip}_{\epsilon} \left(i_t^{adv}+ \delta_t \right) $
\ENDFOR
\ENDFOR
\end{algorithmic}
\label{method}
\end{algorithm}

\subsection{modalities expansion}
In the modalities' expansion, we construct multi-caption containing diverse textual instruction with the same semantic intent. In our proposed framework, we employ Gemini 2.5 ~\cite{gemini2023} and Video-LLVA ~\cite{lin2024videollava} to generate semantically equivalent captions for each video based on time order of the whole driving scenario. After that, we utilizes semantic entropy~\cite{farquhar2024semantic} to filter redundant captions and rank semantically informative descriptions for temporal frame selection using CLIP-based visual semantic alignment. Given an input video $V$, our framework begin to generate set of captions $C = C^{G} \cup C^{L}$ where $C^{G}$ represents captions generated by Gemini and $C^{L}$ denotes to the captions generated by Video-LLAVA. Thus, all captions generated are defined as  $C = \{c_1,c_2,\dots,c_K\}$ where $K$ donate the total number of generated captions. \\
To candidate the best three captions among them, we evaluate the semantic relevance to the visual content of the video. CLIP ~\cite{radford2021clip} is used as a shared vision-language embedding model. Each frame $f_t$ is encoded using the CLIP visual encoder $v_t=\Phi_v^{CLIP}{f_t}$, where each caption $C_K$ is encoded using CLIP text encoder $u_k=\Phi_c^{CLIP}(c_k)$. The visual semantic similarity between frame $f_t$ and caption $c_k$ is computed as equation follow:
\begin{equation}
    \cos(v_t,u_k)=\frac{v_t^\top u_k}{|v_t|_2|u_k|_2}
\end{equation}

Since the generated captions are intended to describe the entire video rather than individual frames, the final relevance score of each caption is obtained by averaging its similarity across all sampled frame:
\begin{equation}
    \frac{1}{T}\sum_{t=1}^{T}\cos(v_t,u_k)
\end{equation}

\begin{figure}
    \centering
    \includegraphics[width=1\linewidth]{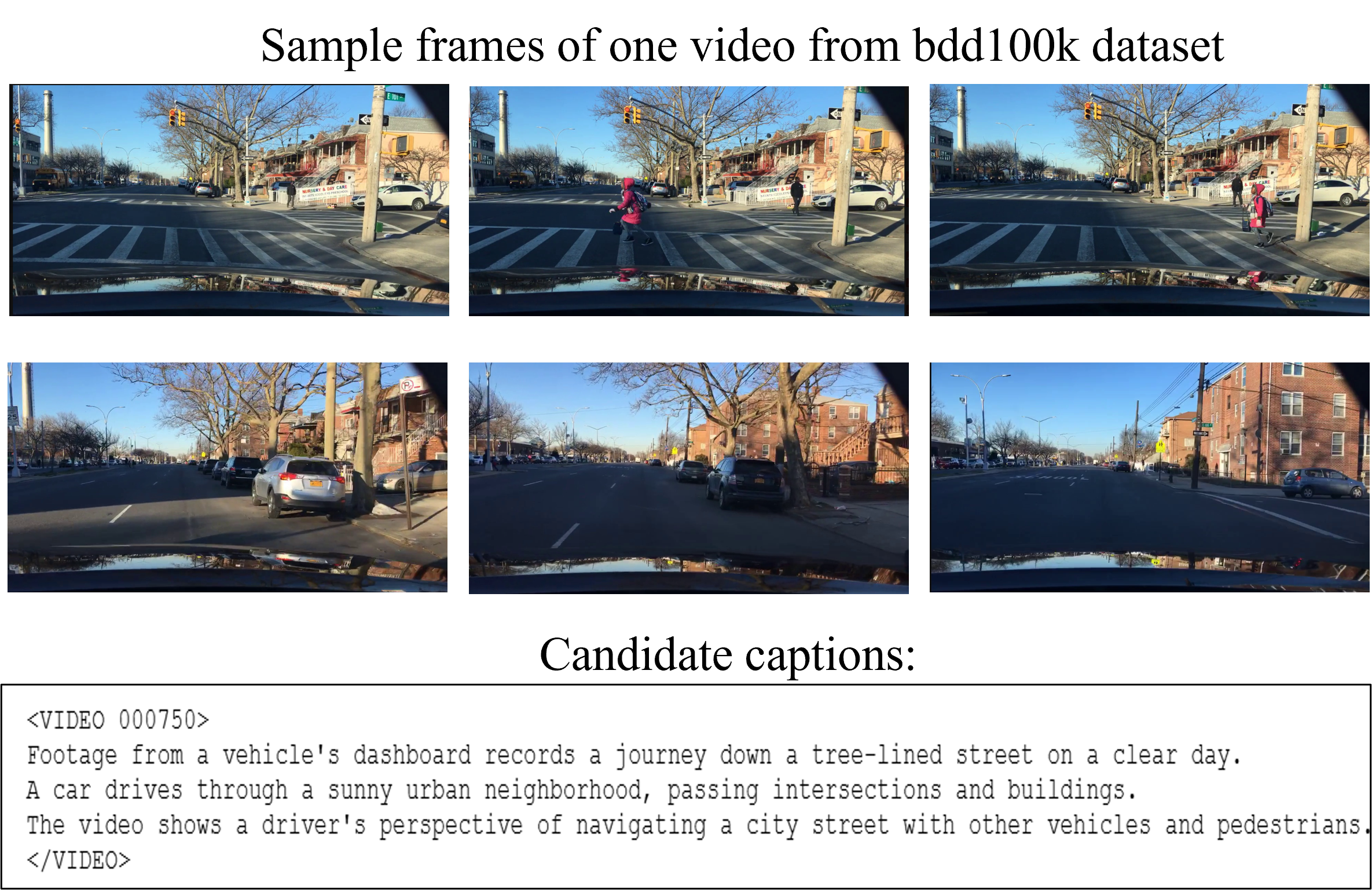}
    \caption{Caption selected strategy}
    \label{fig:caption-sel}
\end{figure}

While caption with higher similarity value indicate stronger semantic correspondence aligned with visual content of the video. Finally, the top-ranked captions are selected as: $C=\operatorname{TopK}\big(s(c_k,v)\big)$. This caption selected strategy enable the framework to retain captions that are strongly grounded in the visual semantics of the input video as shown in figure \ref{fig:caption-sel}.

\subsection{Spatial Attack Stage}
Instead of perturbing all frames uniformly, we introduce a \textbf{caption-guided frame selection} strategy to identifies semantically informative frames using caption-guided visual-semantic alignment. We aim to focus on frames that exhibit stronger semantic correspondence with candidate captions. Given video $V$ consisting of $N$ frames $\{f_1,f_2 \dots f_N\}$ and set of candidate caption $C$ associated with the video. We employ CLIP as a cross-modal feature extractor. For each frame $f_i$, we extract its visual embedding using the CLIP image encoder $E_i$, and for each caption $t_i$, we extract its textual embedding using the CLIP text encoder $E_t$:
\begin{equation}
 v_i = \frac{E_I(f_i)}{||E_I(f_i||_2} , u_i=  \frac{E_T(t_j)}{||E_T(t_j||_2}
\end{equation}

The aggregated text representation is computed as the mean of all caption embeddings: 

\begin{equation}
\bar{u} = \frac{\sum_{j=1}^M u_j}{||\sum_{j=1}^M u_j||_2}
\end{equation}

The semantic similarity between frame $f_i$ and the caption set is measured via cosine similarity $s_i = v_i^\top \bar{u}$. The top K frames are selected based on their similarity scores:
\begin{equation}
    \mathcal{F}^* \leftarrow \underset {\mathcal{S} \subseteq \{1,\dots ,N\},| \mathcal{S}|=k}{\arg \max} \sum_{i \in \mathcal{S}}s_i
\end{equation}

\begin{figure}
    \centering
    \includegraphics[width=1\linewidth]{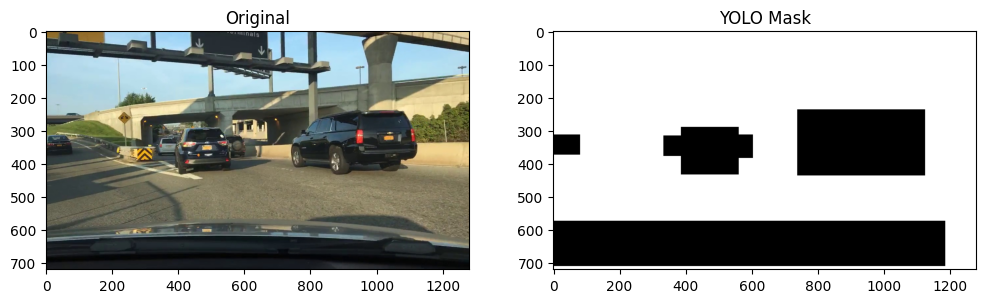}
    \caption{Sample of Mask Generation}
    \label{fig:mask}
\end{figure}

After select vital frames, we applied \textbf{mask} to target specific areas of the selected frames that semantically significant as seen in figure \ref{fig:mask}. Our mask generation method based on YOLO object detector to concentrate the adversarial perturbation to regions containing pivotal objects. Given frame $f_i$, we apply the YOLOv8 detector $\mathcal{D}$ to obtain a set of K bounding boxes. $\mathcal{B}_i = \mathcal{D}(f_i) = \{b_1,b_2, \dots ,b_K \} , b_K =(x_1^k , y_1^k, x_2^k , y_2^k ) $ where each bounding box $b_k$ is retained only if it is confidence score $c_k$ exceeds the threshold $\delta$. The binary precision $M \in \{ 0,1\}^{H \times W}$ is constructed as follows:
\[
\Tilde{M}_{ij}= \begin{cases} 1 , &\text{if } \exists ,b_k \in \mathcal{B}_i : x_1 ^{k} \leq j \leq x_2^{k} \text{ and } y_1 ^{k} \leq j \leq y_2^{k} \\ 0 , &\text {otherwise} \end{cases}   
\]
Where $M_{ij} =1$ are designated as attack zones corresponding to detected object, while regions where $M_{ij} =0$ remain unperturbed. 

For each selected frame $f_i$ and its corresponding mask $M_i$, we generate adversarial example using the Precision Mask Perturbations (PMP) framework ~\cite{fu2024pgattack}.The adversarial perturbation is optimized to maximize the misalignment between the visual and textual representations in the shared embedding space. For each video, we extract the textual embedding of each caption $C_j$, the adversarial optimization is formulated as: 
\begin{equation}
  \ i^{adv} = \underset {||i' -f_i||_{\infty} \leq {\epsilon} }{\arg \max} \mathcal{L} (E_I (i'),\{E_T(C_j)\}_{j=1}^3 \
\end{equation}

Where $i'$ represent adversarial frame, $f_i$ represent original frame, $E_I$ is  image encoder and $E_T$ donate to textual encoder. The maximum perturbation budget should not exceed $\epsilon$ value. $\mathcal{L}$ is the adversarial loss function defined as:
\begin{equation}
    \mathcal{L}= - \sum_{j=3}^{3} (E_I(i'). E_T(C_j))
\end{equation}
The precision mask $M_i$ ensure that perturbations are restricted to semantically meaningful object regions. The multi-scale augmentation strategy applies the perturbation across scales $S =\{ 0.5 , 0.75 , 1.0 ,1.25,1.5\}$ to enhance transferability.
\subsection{Temporal Attack Stage}
While spatial attack successfully disrupts the semantic alignment between individual frames and their corresponding captions, it does not explicitly target the temporal dynamic that VLM model upon for video understanding. To address this limitation, we propose a \textbf{temporal adversarial attack} that degrading the temporal coherence between two consecutive frames, and disrupting the video-text semantic alignment. We hypothesize that by simultaneously corrupting the spatial semantic and temporal coherence of the video, the model's ability to understand and describe driving scenarios can be significantly impaired.\\
The adversarial video produced in spatial attack, is fed into the temporal attack to apply additional temporally-aware perturbations. To focus perturbations on semantically meaningful dynamic regions, we compute motion masks from consecutive masks of two adjacent frames. Given the binary mask $M$, the motion mask between frames $t$ and $t+1$ is defined as: $\mathcal{M}_{motion} = |\mathcal{M}_{t+1} - \mathcal{M}_t| > \mathcal{T}_m $ where $\mathcal{T}_m $ =0.1 is the binarization threshold. This motion masks ensure that perturbation are restricted to regions where object are actively moving, preserving the visual quality of static background region. The total adversarial loss is calculated by \textbf{Text-Visual Alignment Loss} that measure the cosine similarity between the video representation and the textual description embedding:
\begin{equation}
    \mathcal{L}_{text} =\frac{1}{|\mathcal{T}|} \sum_{j=1}^{|\mathcal{T}|} \cos (E_V (i^{adv}), E_T(t_j))
\end{equation}
Where $E_V$ denotes the LanguageBind video encoder, $E_T$ denotes the language encoder, and $\mathcal{T}$ is the set of captions associated with video. Maximizing $\mathcal{L}_{text}$ drive the adversarial video representation away from its corresponding textual description.\\ 

Additionally, we calculate \textbf{Temporal Coherence Loss} which measure the cosine similarity between features of temporally adjacent frame pairs. 
\begin{equation}
    \mathcal{L}_{temporal} = \frac{1}{K-1} \sum_{t=1}^{K-1} \mathcal{w}_t \cdot \cos (E_V (i_t^{adv}), E_V(i_{t+1})) 
\end{equation}

Where K is the total number of motion frames, and consecutive frame pairs are used to capture full temporal coherence across the entire video sequence.

\section{Experiments}
\subsection{Experimental Setup }
\textbf{Dataset.} We evaluate our proposed attack framework on the BDD100K dataset, a large-scale autonomous driving benchmark containing diverse driving scenarios across multiple cities, weather conditions , and times of day. Following our data preparation pipeline, we select 1000 videos in Caption-Guided frame selection strategy, from which 800 videos are used for adversarial attack generation and evaluation. Additionally, we evaluate our approach on nuSenes autonomous driving dataset, we utilize 85 scenes from front-camera which each scene span 20 seconds and contains approximately 40 frames. \\
\textbf{Target Model.} we evaluate our attack against three target video language model with diverse architecture and training paradigms, Video-LLaVA-7B, an open source video language model that align video and image encoder with LLaMA backbone as its language decoder through shared projection mechanism. Second recent model is Qwen2.5-VL-7B which also open source multi-modal that integrate advanced visual perception with deep language understanding, achieving competitive performance in video understanding tasks. The third victim model is Dolphin that domain specific VLM finetuned for autonomous driving scenes and trained on video-text pair driving specific. Dolphins is built upon OpenFlamingo architecture with MPT 7B language backbone.  \\

\textbf{Surrogate Model.} For \textbf{spatial attack} stage, we use TCL as the white box surrogate model with vision transformer ViT-B/16 with 12 transformer layer and 85.8M parameters, built upon the ALBEEF architecture with a ViT-B/16 visual encoder and BERT-base text encoder. For the \textbf{temporal attack} stage, we utilize LanguageBind, an open source model, it align different modalities unless text into  a shared video language embedding space using language as the anchor. it is framework extends CLIP's ViT-L/14 encoder. \\

\textbf{Attack Hyperparameters.} For frame selection, we use CLIP ViT-B/32 to compute cosine similarity between frames and captions, selecting 60 frames per video. Object detection is performed using YOLOv8s with confidence threshold 0.25 and IoU threshold 0.45 attack parameters. For spatial attack, we adopt PGD based optimization with maximum perturbation budget $\epsilon = 32/255$ under the $\ell \infty$-norm constraint , step size $\alpha =1/255$, and number of iterations $\mathbf{T} =20$ with momentum decay $\lambda = 0.9$. Multi-scale augmentation is applied with scales $\mathbf{S = \{0.5,0.75,1.0.1.25,1.5 }\}$. For the temporal attack, we set $\epsilon = 16/255$ , step size $\alpha =2/255$, and number of iterations $\mathbf{T} =20$. Perturbations are applied to regions identified by motion-guided mask derived from consecutive YOLO mask difference to ensure the adversarial attack is concentrated on temporally dynamic regions.\\
\textbf{Implementation Details.} All experiments are conducted on NVIDIA A100 GPU in Google Colab Pro. To reduce memory, video LLaVA and Qwen2.5-VL are loaded in 16 bit and bfloat 16 precision, respectively. Dolphins is evaluated with its official checkpoint and LoRA fine-tuning configuration. To accommodate GPU memory constraints, we evaluate Dolphin in half-precision (float16). For each model, we query each video twice with identical prompt to ensure the divergence is caused by adversarial perturbation only- once on the original frame and other on adversarial perturbed frame.     \\
\textbf{Evaluation Metrics.} We evaluate our attack using two complementary metrics: 
\begin{itemize}
    \item \textbf{Attack Success Rate (ASR):} the percentage of videos for which VLM model generates a description after the attack that differs significantly from its description before the attack. We measure response divergence using word-overlap similarity between pre-attack and post-attack response; when the overlap fall below 0.5, it is considered to successful attack.    
    \item \textbf{Structural Similarity Index (SSIM):} measure perceptual quality of adversarial frames relative to original frames.
\end{itemize}

\subsection{Main Results}
\begin{figure}
    \centering
    \includegraphics[width=1\linewidth]{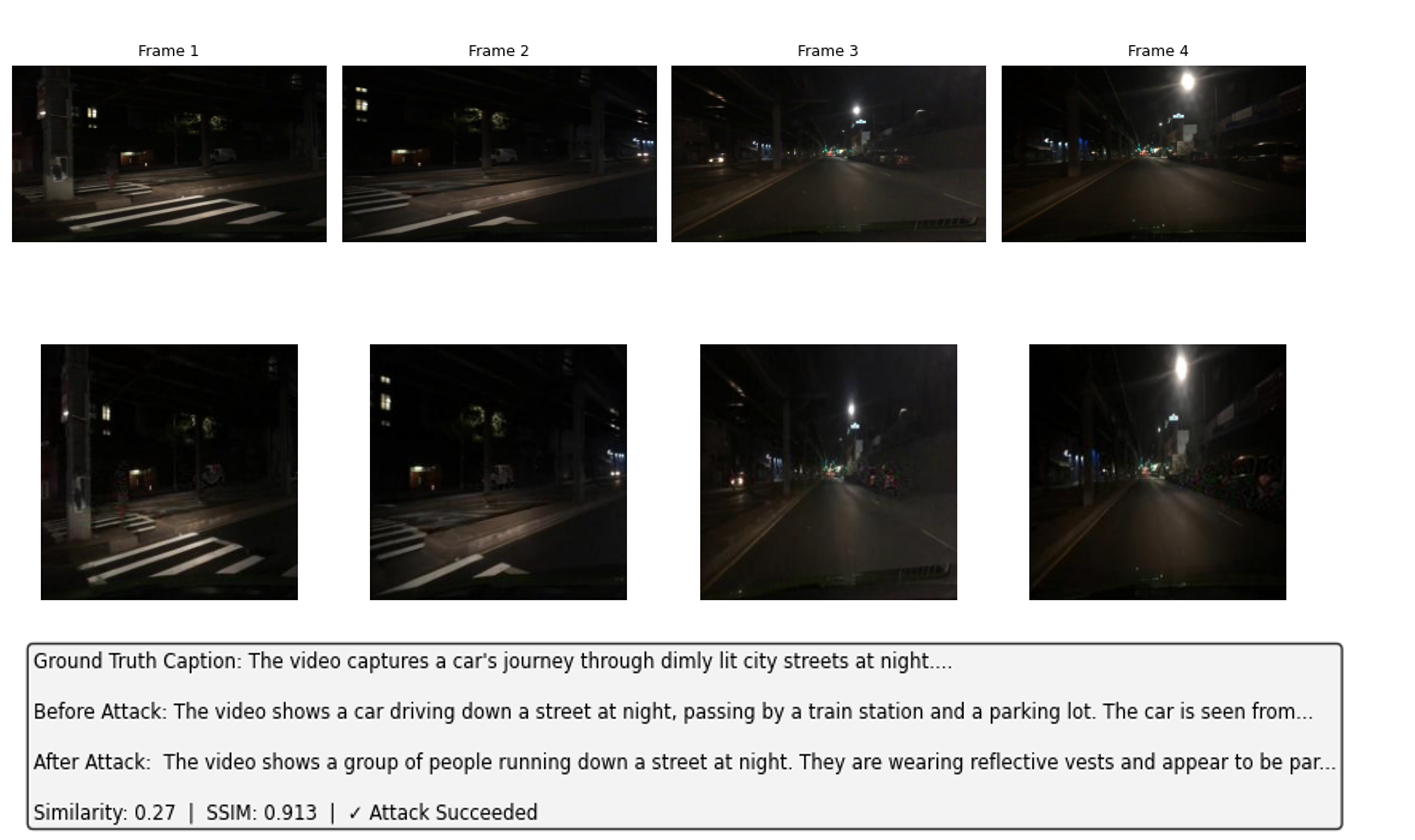}
    \caption{Captions resulted from Video-LLaVA model before and after attack}
    \label{fig:llava-result}
\end{figure}

Performance of our spatial and temporal attacks against all models on 800 driving videos from BDD100K dataset is presented in Table ~\ref{tab:placeholder} and nuScene dataset presented in table ~\ref{tab:placeholder2}, we report the attack success rate and mean SSIM achieved in each stage.\\
Our framework STCA attack consistently outperform the spatial attack only across all target models and both datasets, which confirm our hypothesis that temporal coherence  attack is more effective than spatial attack alone for VLM. The spatial attack alone achieves an attack success rate 32.6 \% with mean SSIM 0.93 against video-LLaVA-7B on BDD 100K showing that spatially perturbation can mislead model in about one-third of cases. Once the temporal attack stage is added, attack success rate arise more than double to 71\% with SSIM 0.82 which remain well above the threshold commonly  adopted in adversarial attack literature. The two failure mode we target: inter-frame (temporal) and intra-frame (spatial) are complementary vulnerabilities, the sample of its result is shown in figure \ref{fig:llava-result}. \\

\begin{figure}
    \centering
    \includegraphics[width=1\linewidth]{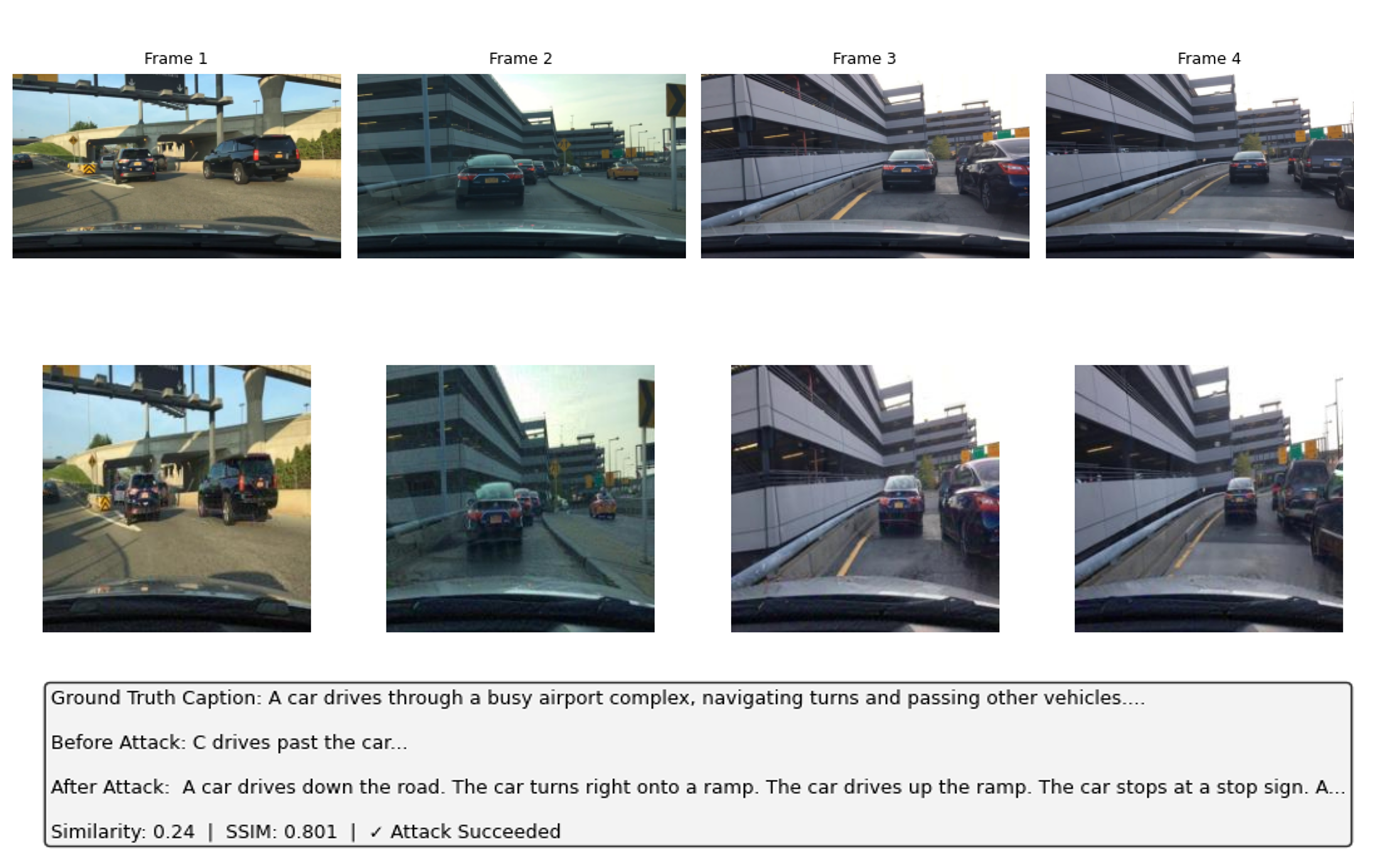}
    \caption{Captions resulted from Qwen2.5-VL model before and after attack}
    \label{fig:Qwen-result}
\end{figure}

\begin{figure}
    \centering
    \includegraphics[width=1\linewidth]{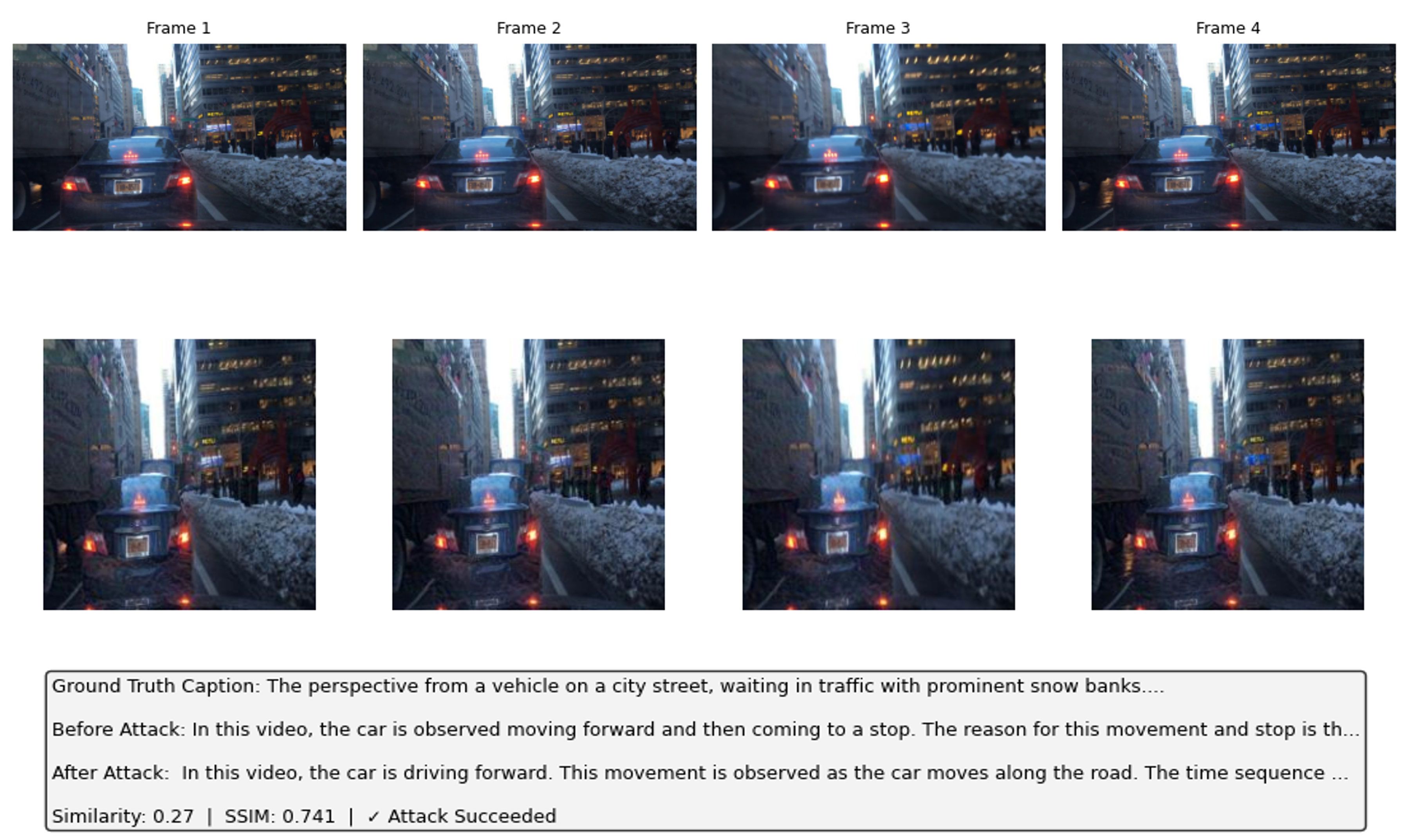}
    \caption{Captions resulted from Dolphins model before and after attack}
    \label{fig:dolphins-result}
\end{figure}

Among the open-source VLM model evaluated in BDD 100K Qwen2.5-VL-7B which proves more susceptible to our STCA attack with attack success rate 84.2\%. Compared to Video-LLaVA's pooling based temporal aggregation, Qwen2.5-VL is more susceptible to localized and motion guided by our attack,the sample of its result is shown in figure \ref{fig:Qwen-result}. By contrast, Dolphins exhibit lower susceptible to our attack, it achieves approximately 47\%  in spatial attack and 46.2\% in our full pipeline. The Dolphins output reveals that, the model has learned driving scene vocabulary as a result of its domain-specific fine-tuning. This finding indicates that domain specialization in VLM may confer partial adversarial resilience,the sample of its result is shown in figure \ref{fig:dolphins-result}. To our knowledge, this observation has not been previously reported for autonomous driving VLM.  \\

\begin{table}[!t]
\caption{Results of our attack in BDD100K dataset}
\centering
\begin{tabular}{|c||c||c|}
\hline
Model & Spatial Attack & Spatial + Temporal Attack \\
\hline
Video LLaVA & 32.6\% & 71\% \\
\hline
Qwen2.5-VL & 45\% & 84.2\% \\
\hline
Dolphin & 46.9\% & 46.2\%\\
\hline
SSIM & 0.93 & 0.82\\
\hline
\end{tabular}
    
\label{tab:placeholder}
\end{table}

\begin{table}[!t]
\caption{Results of our attack in nuScene dataset}
\centering
\begin{tabular}{|c||c||c|}
\hline
Model & Spatial Attack & spatial + Temporal Attack  \\
\hline
Video LLaVA & 57.6\% & 83\% \\
\hline
Qwen2.5-VL & 64.7\% & 96.5\% \\
\hline
Dolphin & 37.6 \% & 47.1\%\\
\hline
SSIM & 0.98 & 0.79\\
\hline
\end{tabular}
    
\label{tab:placeholder2}
\end{table}

\begin{figure}
    \centering
    \includegraphics[width=1\linewidth]{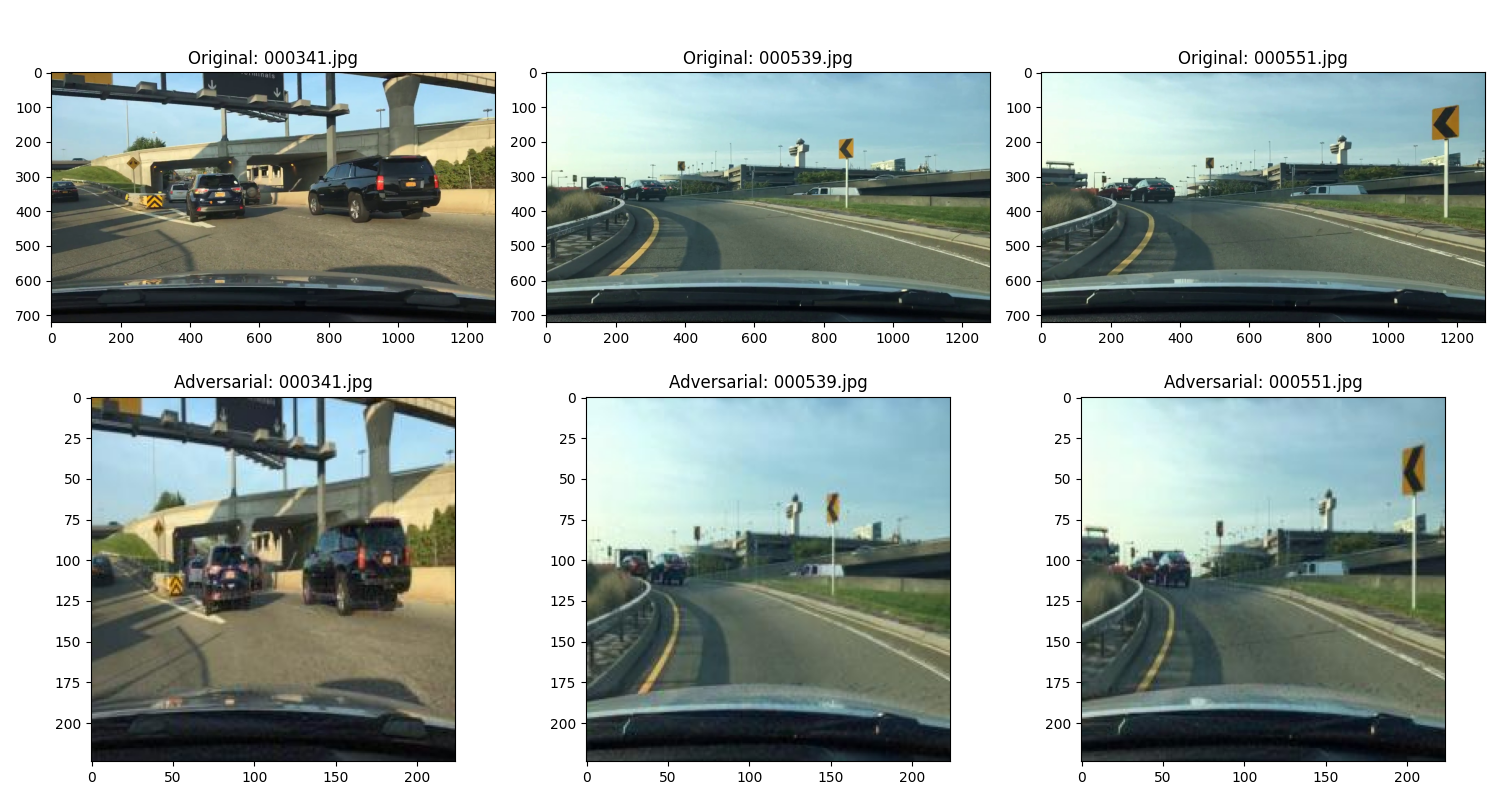}
    \caption{SSIM demonstration between original frames and adversarial frames}
    \label{fig:SSIM}
\end{figure}

On the nuSenes dataset, our spatial attack achieves attack success rate 57.6\%, 64.7\%, and 37\% and achieves in our full pipeline 83\%, 96.5\%, and 47.1\% in video-LLaVA, Qwen2.5-VL, and Dolphins respectively, with mean SSIM 0.89 in spatial temporal and 0.79 in both spatial and temporal stage. Dolphins remains the most robust and Qwen2.5-VL the most susceptible indicating that, the relative ordering of model susceptibility is stable across datasets. The absolute attack success rate on nuScenes are somewhat higher than those on BDD100K due to the shorter temporal extent of nuScenes keyframe sequence. Collectively, the high attack success rate confirm that our proposed temporally-aware attack  framework can deceive various models across autonomous driving benchmark. In term of SSIM value, it remains high which indicate that the adversarial attack of our framework is imperceptible observers as shown in figure \ref{fig:SSIM}. \\

Additionally, we introduce in table \ref{tab:Baseline} two baseline methods, PGD and FGSM and compare them with our framework. Both PGD and FGSM, when applied to the TCL surrogate model under the black box transfer setting, yield substantially low attack success rate than our proposed framework. Specifically, PGD achieves ASR of 47.2\%, 44.8\%, and 36.2\% against Video-LLaVA, Qwen2.5-VL, and Dolphins respectively, while FGSM achieves 25.2\%, 35.4\%, and 18.1\% compared to 71\%, 84.2\% and 46.9\% in our proposed framework. \\
We report in table \ref{tab:results} the most recent studies evaluated in nuSenes dataset. In Video-LLaVa and Qwen2.5 we outperform state of the art methods. In Dolphin model,~\cite{zhang2026advlm} utilized Dolphins in Autonomous driving video and final score drop up 7.49\% and 3.09\%. Also, in other work ~\cite{wang2025black} they outperform prior method by 13.43\%  .For image, Wang et al. ~\cite{wang2025advedm} evaluate dolphins in their AdveDm framework achieving ASR 86\%, this largely ASR attributed to the white box attack setting by leveraging complete access to the target model. Based on our knowledge, there is no previous work evaluate their framework on BDD100K.

\begin{table}[!t]
\caption{Compare our proposed method with baseline methods  }
\centering
\begin{tabular}{|c||c||c||c|}
\hline
Method & Video-LLaVa & Qwen2.5-VL & Dolphins  \\
\hline
PGD & 47.2\% & 44.8\% & 36.2\%\\
\hline
FGSM & 25.2\% & 35.4\% & 18.1\% \\
\hline
Our (Spatial attack only) & 32.6 \% & 45\% & \textbf{46.9\%}\\
\hline
Our (Full pipeline) & \textbf{71\%} & \textbf{84.2\%} & 46.2\%\\
\hline
\end{tabular}
    
\label{tab:Baseline}
\end{table}

\renewcommand{\arraystretch}{1.3}
\begin{table*}[t]
\caption{Comparison of our framework with state of the art methods}
\centering

\label{tab:results}
\begin{tabular}{|c||c||c||c||p{3.5cm}|}
\hline
Dataset & Victim Model & Reference & Attack Type & ASR \% \\
\hline
\multirow{5}{*}{nuScene dataset} & \multirow{4}{*}{Video-LLaVA} & IN et al. (2024) ~\cite{ni2025physical} & Physical backdoor with common object trigger & different based on trigger ranged from 65.3 \% to 89.3\% \\

\cline{3-5}

   & &  Wang et al. (2026)~\cite{wang2025advedm} & Fine grained object-semantic attack  & 75\% \\
\cline{3-5}
   & & \textbf{Spatial Attack only} & Transerable Spatial Attack & 57.6\% \\
\cline{3-5}
   & & \textbf{Spaital+temporal} & Transerable Spatial-Temporal Attack & \textbf{83\%} \\

\cline{2-5}
  & \multirow{2}{*}{Qwen2-VL} & Liu et al. (2025)~\cite{liu2025natural} & Natural-reflection backdoor & 70.92\%\\

\cline{3-5}
 & & \textbf{Spatial Attack only} & Transerable Spatial Attack & 64.7\% \\
\cline{3-5}
 & & \textbf{Spaital+temporal} & Transerable Spatial-Temporal Attack & \textbf{96.5}\% \\

 \cline{1-5}

\end{tabular}
\end{table*}

\subsection{Ablation Study}
\begin{figure*}[!t]
\centering
\subfloat[]{\includegraphics[width=2.5in]{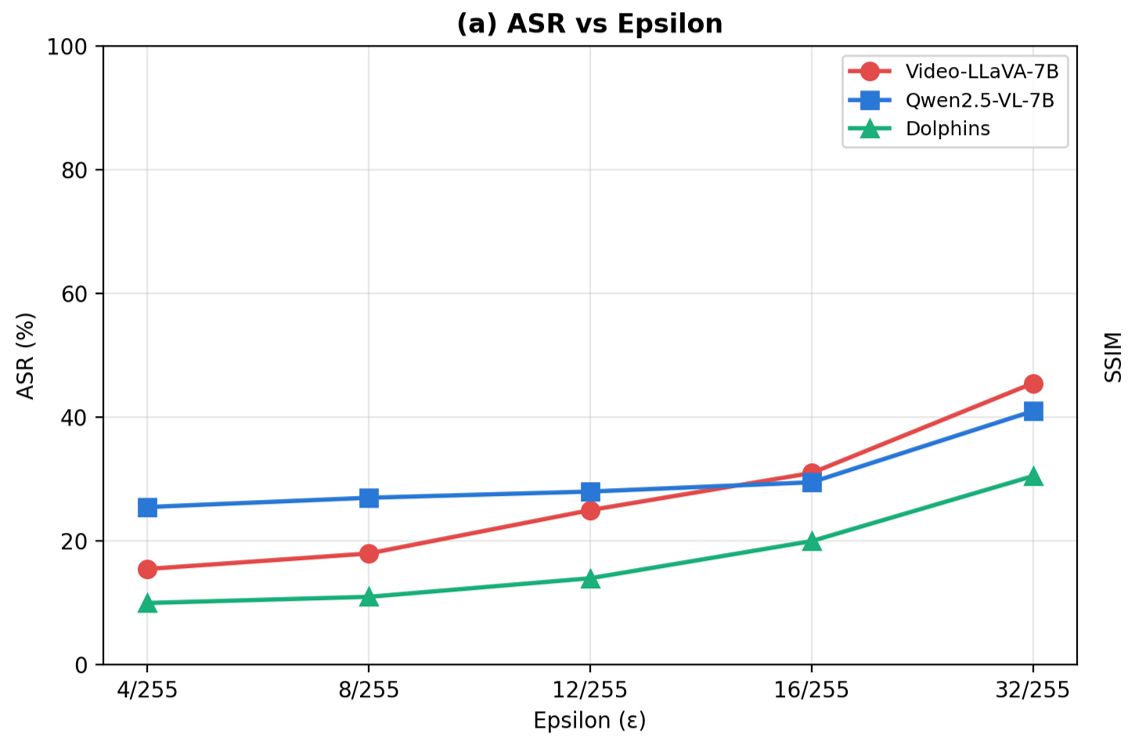}%
\label{fig:eps}}
\hfil
\subfloat[]{\includegraphics[width=2.5in]{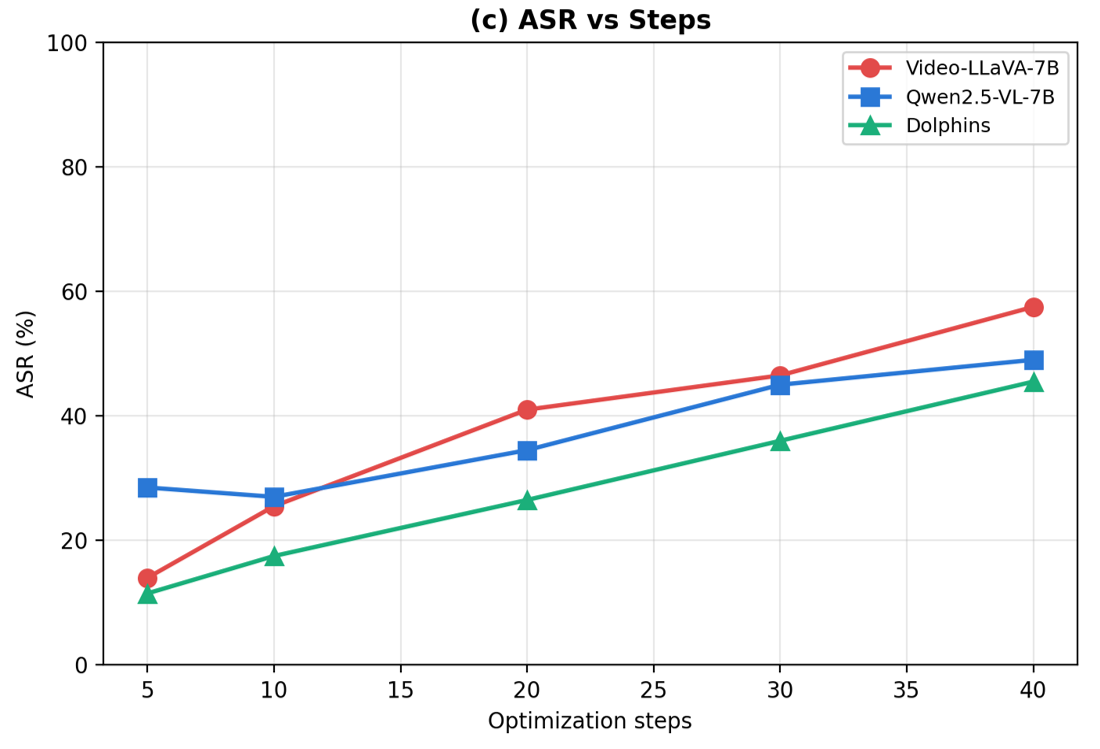}%
\label{fig:step}}
\caption{(a) The relationship between ASR and Perturbation budget. (b) The relationship between the number of ASR and the number of steps.}
\label{fig:ِAblation}
\end{figure*}

We conduct an ablation study to observe the impact of different setting on 200 video from BDD100K. First, we present the result of STCA attack with varying perturbation budgets $\epsilon $ (i.e 4/255 , 8/255, 12/255, 16/255, 32/255) across three model with $n=20$. The specific budgets and results are shown in figure ~\ref{fig:ِAblation} (a), ASR increase consistently with perturbation budget across all three models. Additionally, we tested STCA with different step $n$ (i.e 5, 10, 20, 30, 40) using fixed perturbation budget $\epsilon = 16/255$ and step size $\alpha = 2/255$, generally, the attack strength increase with more iteration step as shown in figure \ref{fig:ِAblation}(b).

\subsection{Discussion And Analysis}
The architecture design of modern vision language model is play a crucial role in achieving higher ASR in our temporal attack than purely spatial attack. These models trained to extract the meaning of video from the temporal relationship between frames, motion pattern, and event sequence that unfold across time. Our attack disrupt precisely the cross frame consistency signal that VLM rely on for temporal reasoning, we achieve this by targeting cross-modal embedding space generated by LanguageBind model with motion guided perturbation applied to temporally dynamic regions. Spatial attack stage disrupting the individual frame's semantics, while temporal stage disrupting a coherence scene description due that, the last delivers substantial gain over the spatial stage only. The two stages complementary target aspects of model's visual understanding.  \\
The key design that contributes to increase both effectiveness and imperceptibility of our attack is the restriction of spatial perturbation to crucial object in each frame and temporal attack to motion- salient regions generated by our YOLO based motion mask computation. Compared to random spatial perturbation strategies, our motion-guided masking achieve higher ration of semantic disruption since moving object already contain variation and temporal inconsistency across frames. \\
The consistent advantage of our attack against Qwen2.5-VL relative to video-LLaVA across both datasets which reflect architectural distinction in how the two model treat temporal information. The substantially lower ASR against Dolphins compared to general purpose open-source models. We finding there is a relationship between model specialization and adversarial robustness so a model with highly tuned for specific domain may exhibit greater robustness to adversarial attack. Additionally, the most practically significant finding of our evaluation, is the successful transfer of our attack from the TCL and LanguageBind surrogates to three architecturally diverse target models. The main factor contribute to this transferability is the motion-guided temporal perturbation proposed in our attack target the temporal coherence of sequential frames.  
\section{Conclusion}
This paper introduce two-stage (STCA) attack targeting video language model in safety-critical issue of autonomous driving. The motivation of our proposed framework is that, the temporal dimension of video input is unexplored in current VLM architectures. Our first stage target spatial object in each frame guided by YOLO-derived mask, followed by the second temporal coherence attack stage target the motion-salient regions between consecutive frames difference. \\
We evaluate our framework across three architecturally divers models: Video-LLaVA, Qwen2.5, and Dolphins, on two large scale autonomous driving benchmarks, BDD100K and nuScenes datasets. Our experimental results demonstrate that, our temporal attack outperforming most recent studies and our spatial-only baseline including PGD and FGSM applied under identical surrogate model.

\vfill

\end{document}